\documentclass[runningheads]{llncs}
\usepackage[T1]{fontenc}
\usepackage{graphicx}
\usepackage{cite}
\usepackage{amsmath,amssymb,amsfonts}
\usepackage{graphicx}
\usepackage{textcomp}
\usepackage{xcolor}
\usepackage{booktabs}
\usepackage{algorithm}
\usepackage{multirow}
\usepackage{array}
\usepackage{dblfloatfix}
\usepackage{float}
\usepackage{cite}
\usepackage{subcaption}
\usepackage{lipsum}
\usepackage{multicol}

\usepackage{algcompatible}
\usepackage{svg}
\usepackage{caption}
\usepackage{graphicx}
\usepackage{afterpage}

\usepackage[colorlinks,citecolor=blue,linkcolor=blue]{hyperref}
\algnewcommand\algorithmicinput{\textbf{Input:}}
\algnewcommand\INPUT{\item[\algorithmicinput]}

\begin{document}

\title{Incorporating Rather Than Eliminating: Achieving Fairness for Skin Disease Diagnosis Through Group-Specific Experts%
\thanks{This is the submitted version of a paper accepted to MICCAI 2025. The final version will appear in the LNCS proceedings published by Springer.}
}
\titlerunning{Achieving Fairness Through Group-Specific Experts}


\author{%
  Gelei Xu\inst{1} \and
  Yuying Duan\inst{1} \and
  Zheyuan Liu\inst{1} \and
  Xueyang Li\inst{1} \and
  Meng Jiang\inst{1} \and
  Michael Lemmon\inst{1} \and
  Wei Jin\inst{2} \and
  Yiyu Shi\inst{1}
}
\authorrunning{Xu et al.}

\institute{%
  University of Notre Dame, Notre Dame, IN 46556, USA \\
  \email{\{gxu4, yduan2, zliu29, xli34, mjiang2, lemmon, yshi4\}@nd.edu}
  \and
  Emory University, Atlanta, GA 30322, USA \\
  \email{wei.jin@emory.edu}
}

\maketitle            
\begin{abstract}
AI-based systems have achieved high accuracy in skin disease diagnostics but often exhibit biases across demographic groups, leading to inequitable healthcare outcomes and diminished patient trust. Most existing bias mitigation methods attempt to eliminate the correlation between sensitive attributes and diagnostic prediction, but those methods often degrade performance due to the lost of clinically relevant diagnostic cues.
In this work, we propose an alternative approach that incorporates sensitive attributes to achieve fairness. We introduce FairMoE, a framework that employs layer-wise mixture-of-experts modules to serve as group-specific learners. Unlike traditional methods that rigidly assign data based on group labels, FairMoE dynamically routes data to the most suitable expert, making it particularly effective for handling cases near group boundaries.
Experimental results show that, unlike previous fairness approaches that reduce performance, FairMoE achieves substantial accuracy improvements while preserving comparable fairness metrics.
\end{abstract}

\section{Introduction}

Recently, AI-based systems have been widely adopted for skin disease diagnostics, achieving high accuracy in predicting various conditions~\cite{ahammed2022machine, Kam_Is_MICCAI2024,zhou2022background}. However, these systems often exhibit biases across population groups. For example, a model may perform well overall but yield inconsistent diagnostic accuracy between patients with dark and light skin types~\cite{groh2021evaluating}. In clinical settings, such biases can lead to inequitable healthcare outcomes and erode patient trust in AI-driven diagnostics, highlighting the urgent need for fairer diagnostic methods.

The most common bias mitigation strategies focus on \textbf{removing} the influence of sensitive attributes (e.g. age, gender). Adversarial training, for example, achieves fairness by training an adversary network to predict sensitive attributes from learned representations while optimizing the main model to minimize this adversary’s success, ensuring that sensitive information is not encoded~\cite{alvi2018turning,zhang2018mitigating}. Regularization-based methods learn fair model by penalizing correlations between sensitive attributes and model's output~\cite{jung2021fair,quadrianto2019discovering}. More recently, pruning and quantization have been used to improve fairness by removing model components that contribute most to disparities across sensitive attributes~\cite{chiu2023toward,wu2022fairprune,guo2024fairquantize}.

While these approaches promote fairness by suppressing sensitive attributes, they may not be ideal for skin disease diagnosis, where such attributes hold essential clinical value. 
For example, skin type encodes diagnostic signals such as UV susceptibility, which is crucial for clinical decision-making~\cite{caini2009meta}. Suppressing these attributes can inadvertently discard vital information, leading to significant declines in model performance~\cite{zafar2019fairness,wang2022understanding}. More concerningly, reducing performance gaps between groups is often achieved not by improving disadvantaged groups but by uniformly degrading performance across all groups~\cite{wu2022fairprune, duan2024post}. Thus, instead of removing sensitive attributes, this paper advocates for \textbf{incorporating} group information to improve fairness. Given the differences in disease distribution across groups, a natural approach is to assign each group its own module to better capture group-specific patterns and enhance diagnostic accuracy.

To achieve this, we draw inspiration from recent advances in Mixture of Experts (MoE), a framework that enables the training of specialized experts with dynamic input routing~\cite{shazeer2017outrageously,zhang2023robust}. MoE provides a promising solution for group-aware learning by allowing models to leverage different experts based on input characteristics. 
However, directly applying MoE poses several challenges. \textbf{First}, MoE routers automatically learn expert selection, making it difficult to explicitly assign experts to specific groups. \textbf{Second}, balancing specialization and generalization is non-trivial. Training an expert solely on its group's data strengthens specialization but weakens generalization due to data scarcity, while incorporating other groups' data improves generalization but introduces distribution shifts, reducing specialization. \textbf{Third}, handling continuous attributes (e.g., age) is challenging, especially for intermediate samples that lack clear group assignment. If each expert is optimized for a specific group, borderline cases may experience degraded performance due to ineffective expert coverage. Addressing these challenges is crucial for using MoE in fairness-aware medical AI systems.

Building on these insights, we introduce FairMoE, a framework that employs layer-wise mixture-of-experts modules to serve as group-specific learners for skin disease diagnosis. FairMoE leverages mutual information loss to establish strong correlation between groups and experts, ensuring that each expert specializes in a specific group.
To maximize performance across all groups, particularly for data samples that fall around group boundaries, each expert is trained not only on its designated group's subset but also selectively incorporates data from other groups through soft probabilistic routing. Unlike previous approaches that often degrade the performance of advantaged groups, FairMoE improves all group's performance while maintaining comparable fairness metrics. 


\section{Motivation}\label{background}

In applications such as recidivism risk assessment~\cite{flores2016false} or income prediction~\cite{asuncion2007uci}, fairness requires excluding sensitive attributes such as race and gender from training to prevent bias. However, in skin disease diagnosis, these sensitive attributes provide essential clinical insights and cannot simply be removed.
For example, skin type offers crucial information for clinicians assessing UV susceptibility~\cite{caini2009meta}, while race and gender also influence disease prevalence—malignant melanoma is 20 times more common in African Americans compared to other groups~\cite{narayanan2010ultraviolet}. Similarly, men are twice as likely to develop basal cell carcinoma (BCC) and three times more likely to develop squamous cell carcinoma (SCC) than women~\cite{gordon2013skin}.
Therefore, removing sensitive attributes as done in most fairness-aware algorithms risks compromising performance, which is not an option in medical diagnostic models where delivering precise diagnoses is absolutely crucial. While existing approaches~\cite{wang2020mitigating, puyol2021fairness} attempt to improve fairness by explicitly incorporating sensitive attributes and training separate models for different demographic groups by only using in-group data, such models often suffer from limited generalization due to reduced training data. Motivated by the needs, this paper proposes FairMoE, which integrates sensitive attributes to enhance fairness while training group-specific experts. Unlike methods mentioned above, FairMoE selectively incorporates out-of-group data to improve generalization, achieving fairness while maintaining model's performance.

\section{Methodology}




\subsection{Problem Definition and Introduction of MoE Model}


\noindent\textbf{Problem Definition.} 
Consider a dataset \( D = \{(x_i, y_i, c_i)\}_{i=1}^N \), where \( x_i \in \mathcal{X}\) represents the input image, \( y_i \in \mathcal{Y}\) is the ground-truth class label, and \( c_i \in \mathcal{C} = \{C_1, C_2, \dots, C_m\} \) denotes the sensitive attribute (e.g., age, skin type), which we also refer to as ``group label'' in this paper.
We define our MoE model as \( \hat{y}_i = F(\theta, x_i) \), where \( F \) maps the input \( x_i \) to the predicted label \( \hat{y}_i \) and is parameterized by weights \( \theta \).  Our objective is to enhance model performance across different groups by leveraging the sensitive attributes, we incorporate  \( c_i \) during training to guide the routing mechanism. Accordingly, we denote the MoE model as  
\(
\hat{y}_i = F(\theta, x_i, c_i),
\)  
explicitly indicating that \( c_i \) is included in the training.

\noindent\textbf{MoE Model.} Our MoE model extends a convolutional neural network (CNN) backbone by replacing each convolutional (Conv) layer with a MoE Conv Layer. A MoE Conv Layer comprises a set of expert networks \(E_1, E_2, \dots, E_n\) and a routing network \(G\)~\cite{shazeer2017outrageously}. Each expert is a Conv layer with the same dimensions as the corresponding Conv layer in the original backbone CNN. The routing network is a significantly smaller Conv layer that directs data flow.  
Each expert is specialized for a specific group, so we set the number of experts \(n\) equal to the number of groups \(m\). During training and inference, the routing network dynamically selects an expert for a given sample and forwards it to the next layer. The core of FairMoE lies in the routing network, which governs how experts receive training data and determines expert selection during inference.

\begin{figure}[t]
    \centering
    \includegraphics[width=1\textwidth]{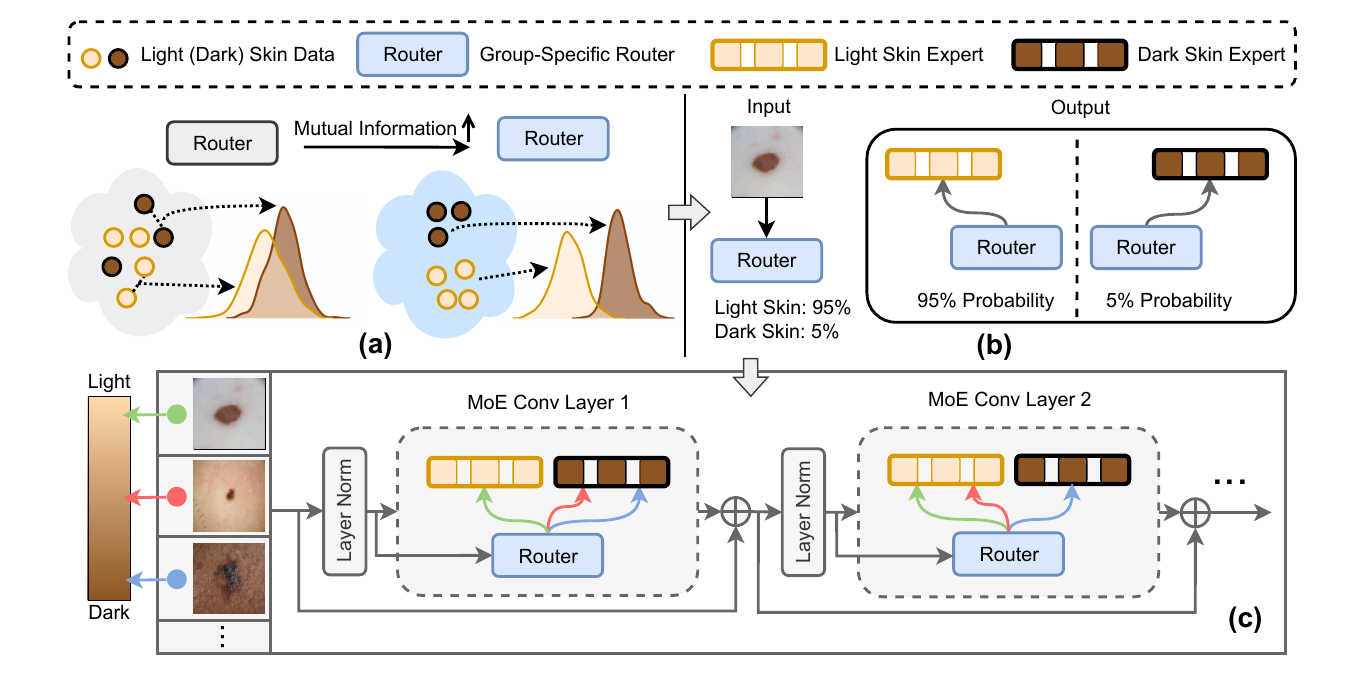}
    \caption{Overview of FairMoE Framework: (a) Establishing correlations between groups and experts, (b) implementing soft probabilistic expert routing, and (c) illustrating the expert selection process within a ResNet-18 module. }
    \label{fig:overview}
\end{figure}

\subsection{FairMoE Framework Overview}

Figure~\ref{fig:overview} provides an overview of the FairMoE framework, where the sensitive attribute is binary skin tone, and each layer has two corresponding experts. The design of FairMoE consists of two key components: First, as shown in Figure~\ref{fig:overview}a, the router maps each sample to the expert corresponding to its group. To ensure that samples from a given group are forwarded to their group-specific expert with high probability, the router maximizes the mutual information between the group label $C_i$ and the assigned expert $E_i$. Second, as shown in Figure~\ref{fig:overview}b, instead of rigid assignments, the router uses a probabilistic selection strategy to mitigate data scarcity in group-specific training, particularly benefiting boundary samples. With both components successfully deployed, Figure~\ref{fig:overview}c illustrates the expert selection process within a ResNet-18 with two MoE Conv Layers. Samples for which the router has high confidence in their group membership are assigned to the corresponding expert with high probability. While for samples near the decision boundary—such as those with intermediate skin tones—expert selection becomes more stochastic, leading to dynamic assignments across different layers.

\subsection{Encouraging Specialization Through Group-Aware Assignments
}


In MoE training, when the router's assignment is not constrained by group information, samples are assigned to experts with nearly uniform probability, preventing the formation of group-specific expertise. However, our approach seeks to specialize experts for distinct groups, ensuring each expert is optimized for classifying data within its respective group.

To quantitatively enforce this expert-group specialization, we define $P(E_i|C_j)$ as the probability of a sample from group $C_j$ being assigned to expert $E_i$. 
Using conditional probability, the joint probability is $P(E, C)=P(E|C)P(C)$, where $P(C)$ is a constant depending on the relative sizes of different groups in the dataset.
In order to explicitly build dependence between experts and groups, we measure their correlation using mutual information, formulated as:
\begin{equation}
    I(C; E) = \sum_{i=1}^{m} \sum_{j=1}^{m} P(C_i, E_j) \log \frac{P(C_i, E_j)}{P(C_i)P(E_j)}.
\end{equation}
In this context, \( m \) represents both the number of experts and the number of groups. If experts are assigned to all groups with equal probability, meaning there is no correlation between groups and experts, then the joint probability satisfies $P(C_i, E_j) = P(C_i) P(E_j)$, resulting in a mutual information value of zero. Conversely, if each expert is exclusively assigned to a single group, the mutual information is maximized. In practice, we incorporate \(-I(C; E_Y)\) into the total loss for each MoE layer \( Y \), weighted by a parameter \( w_{MI} \), where \( E_Y \) represents all experts within layer \( Y \).



\subsection{Soft Probabilistic Expert Routing}

If a strong dependency forms between experts and groups, the router effectively functions as a group classifier, consistently assigning each sample to a fixed expert. While this structure enables expert specialization, it introduces two major limitations that ultimately compromise model performance. First, confining an expert to a single group reduces its available training data, weakening generalization, especially in imbalanced datasets. For example, if a group contains only 10\% of the dataset, its expert trains on a limited subset, restricting its ability to learn robust patterns and resulting in suboptimal performance. Second, for continuous attributes like skin tone or age, hard group assignments fail to capture boundary features effectively. For instance, when groups are defined as age $<$ 55 and age $\ge$ 55, experts trained exclusively on these subsets struggle to generalize in transition regions (e.g., ages 50–60), leading to reduced predictive accuracy.

To balance specialization and generalization while enhancing performance on group boundary samples, it is crucial to selectively incorporate out-of-group data when training group-specific experts. To minimize distribution shift, priority should be given to samples closer to the target group’s distribution.
We achieve this by leveraging the router’s confidence score as a proxy for a sample’s proximity to each group’s distribution, using it as the probability for expert selection. For example, a sample with a high probability of belonging to the light-skin group should primarily train the light-skin expert, whereas assigning it to the dark-skin expert would introduce distribution shift. Conversely, a sample with equal probabilities for both groups should contribute equally to both experts. This approach replaces the rigid, hard-assignment strategy—where each sample is strictly assigned to the expert based on its predicted group—with a soft, probabilistic assignment, where selection probability reflects a sample’s contribution to different groups.
The probability for selecting expert $E_k$ is given by:

\begin{equation}
P(E_k \mid x) = \frac{\alpha_k \cdot s_k(x)}{\sum_{j} \alpha_j \cdot s_j(x)},
\end{equation}
where \( s_k(x) \) denotes the router’s confidence score for assigning sample \( x \) to group \( k \), and \( \alpha_k = \frac{1}{N_k} \) balances expert selection based on group sizes \( N_k \). Without $\alpha_k$, the router would disproportionately assign samples to majority-group experts, leading to uneven training.


\section{Experiments and Results}\label{experiment}


\noindent\textbf{Dataset and Model.} FairMoE is evaluated on two skin disease classification datasets: Fitzpatrick-17k~\cite{groh2021evaluating} and ISIC 2019~\cite{combalia2019bcn20000,tschandl2018ham10000}. Fitzpatrick-17k contains 16,577 images representing 114 different skin conditions, with skin types 1-3 grouped as light skin and 4-6 as dark skin, following~\cite{wu2022fairprune}. ISIC 2019 contains 25331 images across 8 diagnostic categories, where age as the sensitive attribute, grouping individual $\le 55$ as young and $\textgreater 55$ as old. We use ResNet-18~\cite{he2016deep} as the backbone for Fitzpatrick-17k and VGG-11~\cite{simonyan2014very} for ISIC 2019, extending their original convolutional layers to MoE Conv layers. The training settings were the same with those described in~\cite{wu2022fairprune}. A standard preprocessing step for both datasets involves resizing all the images to a uniform size of 128×128 pixels. The mutual information loss constant \( w_{MI} \) is set to 0.01.


\noindent\textbf{Fairness Metrics.}
We employ the multi-class equalized opportunity (Eopp)
and equalized odds (Eodd) metrics to evaluate the fairness of our model.
We follow~\cite{wu2022fairprune} for calculating these metrics. We use the FATE metric from~\cite{xu2023fairadabn} to assess the overall performance of the model balancing accuracy and fairness. A higher FATE score indicates a better trade-off between fairness and accuracy. 

\noindent\textbf{Baselines.}
We present a comprehensive comparison of our framework against various bias mitigation baselines. Vanilla is the model trained directly on ResNet-18 or VGG-11 backbone. DomainIndep learns group-specific classifiers with shared parameters~\cite{wang2020towards}. FairAdaBN achieves fairness by modifying the batch normalization layer~\cite{xu2023fairadabn}. SCP-FairPrune~\cite{kong2024achieving} and FairQuantize~\cite{guo2024fairquantize} is the latest work achieves fairness on skin disease through pruning and quantization, respectively. 

\noindent\textbf{Results on Fitzpatrick-17k Dataset.}
The upper section of Table~\ref{table:combined} presents a comparative analysis of accuracy and fairness across methods on the Fitzpatrick-17k dataset, highlighting FairMoE's superior performance. FairMoE achieves the highest F1-score (0.502), outperforming the second-best baseline SCP-FairPrune (0.476). Additionally, it attains the lowest Eopp1 and Eodd scores, indicating improved fairness. Furthermore, FairMoE substantially increases FATE scores, with a 46\% increase in FATE (Eopp0) and 104\% in FATE (Eodd) compared to SCP-FairPrune, demonstrating the effectiveness of FairMoE.

\afterpage{
\begin{table}[H]
    \centering
    \caption{Results of accuracy and fairness on Fitzpatrick-17k and ISIC2019 datasets. The best performance is highlighted in bold.}
    \label{table:combined}
    \scriptsize
    \renewcommand{\arraystretch}{0.8}
    \resizebox{\columnwidth}{!}{
        \begin{tabular}{lccccccc}
            \toprule
            & & \multicolumn{3}{c}{Accuracy}  & \multicolumn{3}{c}{Fairness}                    \\ \cmidrule(lr){3-5}\cmidrule(lr){6-8}
            Method &  Group     & Precision & Recall & F1-score & Eopp0 $\downarrow$ / FATE $\uparrow$  & Eopp1 $\downarrow$ / FATE $\uparrow$   & Eodd $\downarrow$ / FATE $\uparrow$   \\ 
            
            \midrule[1.2pt] 
            \multicolumn{8}{c}{\textbf{Fitzpatrick-17k Dataset}} \\ 
            \midrule

\multirow{4}{*}{ResNet-18} & Dark & 0.512& 0.511& 0.490& \multirow{4}{*}{0.0031 / 0.000}& \multirow{4}{*}{0.332 / 0.000}& \multirow{4}{*}{0.180 / 0.000}\\
             & Light &0.467 & 0.468& 0.449& & & \\
             & Avg. $\uparrow$ &0.489 & 0.490& 0.468& & & \\
             & Diff. $\downarrow$ & 0.045& 0.043& 0.041& & & \\
             \midrule
             \multirow{4}{*}{DomainIndep} & Dark & 0.501& 0.522& 0.492& \multirow{4}{*}{0.0030 / 0.041}& \multirow{4}{*}{0.320 / 0.045}&\multirow{4}{*}{0.163 / 0.103} \\
             & Light &0.465 & 0.479& 0.459& & & \\
             & Avg. $\uparrow$ & 0.484& 0.497& 0.472& & & \\
             & Diff. $\downarrow$ &0.036 &0.043 &0.033 & & & \\
             \midrule
             \multirow{4}{*}{FairAdaBN} & Dark &0.493 &0.495 &0.469 & \multirow{4}{*}{0.0030 / 0.007}& \multirow{4}{*}{0.302 / 0.065}& \multirow{4}{*}{0.171 / 0.024}\\
             & Light &0.450 & 0.443& 0.435& & & \\
             & Avg. $\uparrow$ &0.471 & 0.473&0.456 & & & \\
             & Diff. $\downarrow$ &0.043 &0.052 &0.034 & & & \\
             \midrule
             \multirow{4}{*}{SCP-FairPrune} & Dark &0.512 & 0.512& 0.490& \multirow{4}{*}{0.0030 / 0.049}& \multirow{4}{*}{0.289 / 0.147}& \multirow{4}{*}{0.164 / 0.106}\\
             & Light & 0.490& 0.472& 0.469& & & \\
             & Avg. $\uparrow$ & 0.495&0.497 &0.476 & & & \\
             & Diff. $\downarrow$ &\textbf{0.022} &\textbf{0.040} &\textbf{0.021} & & & \\
             \midrule
            \multirow{4}{*}{FairQuantize} & Dark &0.498 &0.513 &0.480 &\multirow{4}{*}{\textbf{0.0028} / 0.082} &\multirow{4}{*}{0.291 / 0.109} &\multirow{4}{*}{0.156 / 0.118} \\
             & Light &0.459 &0.470 &0.448 & & & \\
              & Avg. $\uparrow$ &0.479 &0.489 &0.461 & & & \\
             & Diff. $\downarrow$ &0.043 &0.053 &0.032 & & & \\
             \midrule
            \multirow{4}{*}{FairMoE} & Dark & \textbf{0.540} & \textbf{0.554} & \textbf{0.522} & \multirow{4}{*}{0.0030 / \textbf{0.105}} & \multirow{4}{*}{\textbf{0.285} / \textbf{0.214}} & \multirow{4}{*}{\textbf{0.154} / \textbf{0.217}} \\
    & Light & \textbf{0.505} & \textbf{0.497} & \textbf{0.486} & & & \\
    & Avg. $\uparrow$ & \textbf{0.518} & \textbf{0.525} & \textbf{0.502} & & & \\
    & Diff. $\downarrow$ & 0.035 & 0.057 & 0.036 & & & \\

            \midrule[1.2pt] 
            \multicolumn{8}{c}{\textbf{ISIC2019 Dataset}} \\ 
            \midrule

\multirow{4}{*}{VGG-11} & Young & 0.669 & 0.724 & 0.687 & \multirow{4}{*}{0.023 / 0.000} & \multirow{4}{*}{0.150 / 0.000 } & \multirow{4}{*}{0.078 / 0.000}\\
             & Old & 0.758 & 0.798 & 0.766 & & & \\
             & Avg. $\uparrow$ & 0.723 & 0.773 & 0.741 & & & \\
             & Diff. $\downarrow$& 0.089 & 0.074 & 0.080 & & & \\
             \midrule
             \multirow{4}{*}{DomainIndep} & Young & 0.677 & 0.724 & 0.689 & \multirow{4}{*}{0.021 / 0.090} & \multirow{4}{*}{0.103 / 0.316} & \multirow{4}{*}{0.051 / 0.345} \\
             & Old & 0.758 & 0.796 & 0.766 & & & \\
             & Avg. $\uparrow$ & 0.724 & 0.772 & 0.743 & & & \\
             & Diff. $\downarrow$ & 0.081 & 0.072 & 0.078 & & & \\
             \midrule
             \multirow{4}{*}{FairAdaBN} & Young & 0.643 & 0.717 & 0.663 & \multirow{4}{*}{0.018 / 0.196} & \multirow{4}{*}{\textbf{0.084} / 0.418} & \multirow{4}{*}{0.060 / 0.209}\\
             & Old & 0.727 & 0.770 & 0.748 & & & \\
             & Avg. $\uparrow$ & 0.699 & 0.762 & 0.725 & & & \\
             & Diff. $\downarrow$ & 0.085 & 0.053 & 0.086 & & & \\
             \midrule
             \multirow{4}{*}{SCP-FairPrune} & Young & 0.662 & 0.724 & 0.683 & \multirow{4}{*}{0.020 / 0.125} & \multirow{4}{*}{0.098 / 0.341} & \multirow{4}{*}{0.068 / 0.123}\\
             & Old & 0.745 & 0.792 & 0.761 & & & \\
             & Avg. $\uparrow$ & 0.717 & 0.769 & 0.737 & & & \\
             & Diff. $\downarrow$ & 0.083 & 0.068 & 0.078 & & & \\
             \midrule
            \multirow{4}{*}{FairQuantize} & Young & 0.657 & 0.718 & 0.676 & \multirow{4}{*}{0.019 / 0.163} & \multirow{4}{*}{0.088 / 0.403} & \multirow{4}{*}{0.060 / 0.220} \\
             & Old & 0.743 & 0.785 & 0.755 & & & \\
              & Avg. $\uparrow$ & 0.707 & 0.765 & 0.733 & & & \\
             & Diff. $\downarrow$ & 0.086 & 0.067 & 0.080 & & & \\
             \midrule
\multirow{4}{*}{FairMoE} & Young & \textbf{0.711} & \textbf{0.768} & \textbf{0.725} & \multirow{4}{*}{\textbf{0.016} / \textbf{0.340}} & \multirow{4}{*}{0.089 / \textbf{0.441}} & \multirow{4}{*}{\textbf{0.046} / \textbf{0.445}} \\
    & Old & \textbf{0.782} & \textbf{0.819} & \textbf{0.788} & & & \\
    & Avg. $\uparrow$ & \textbf{0.760} & \textbf{0.795} & \textbf{0.767} & & & \\
    & Diff. $\downarrow$ & \textbf{0.072} & \textbf{0.051} & \textbf{0.063} & & & \\

            \bottomrule
    \end{tabular}}
    \label{table:combined}
\end{table}
}

\noindent\textbf{Results on ISIC 2019 Dataset.}
Similar to Fitzpatrick-17k, FairMoE outperforms all baselines in accuracy and two fairness metrics. Unlike most methods that sacrifice accuracy for fairness, FairMoE effectively avoids this trade-off by employing group-specific experts, allowing each group to optimize its own performance rather than relying on a shared model structure.
Compared to DomainIndep, which also utilizes group-specific modules for fairnessis, FairMoE achieves significantly better accuracy and fairness scores. Additionally, FairMoE exhibits the smallest gap across all accuracy metrics, indicating greater benefits for unprivileged groups and further enhancing fairness.

\begin{figure}[t]
    \centering
    \begin{subfigure}[t]{0.48\textwidth}
        \centering
        \includegraphics[width=\linewidth]{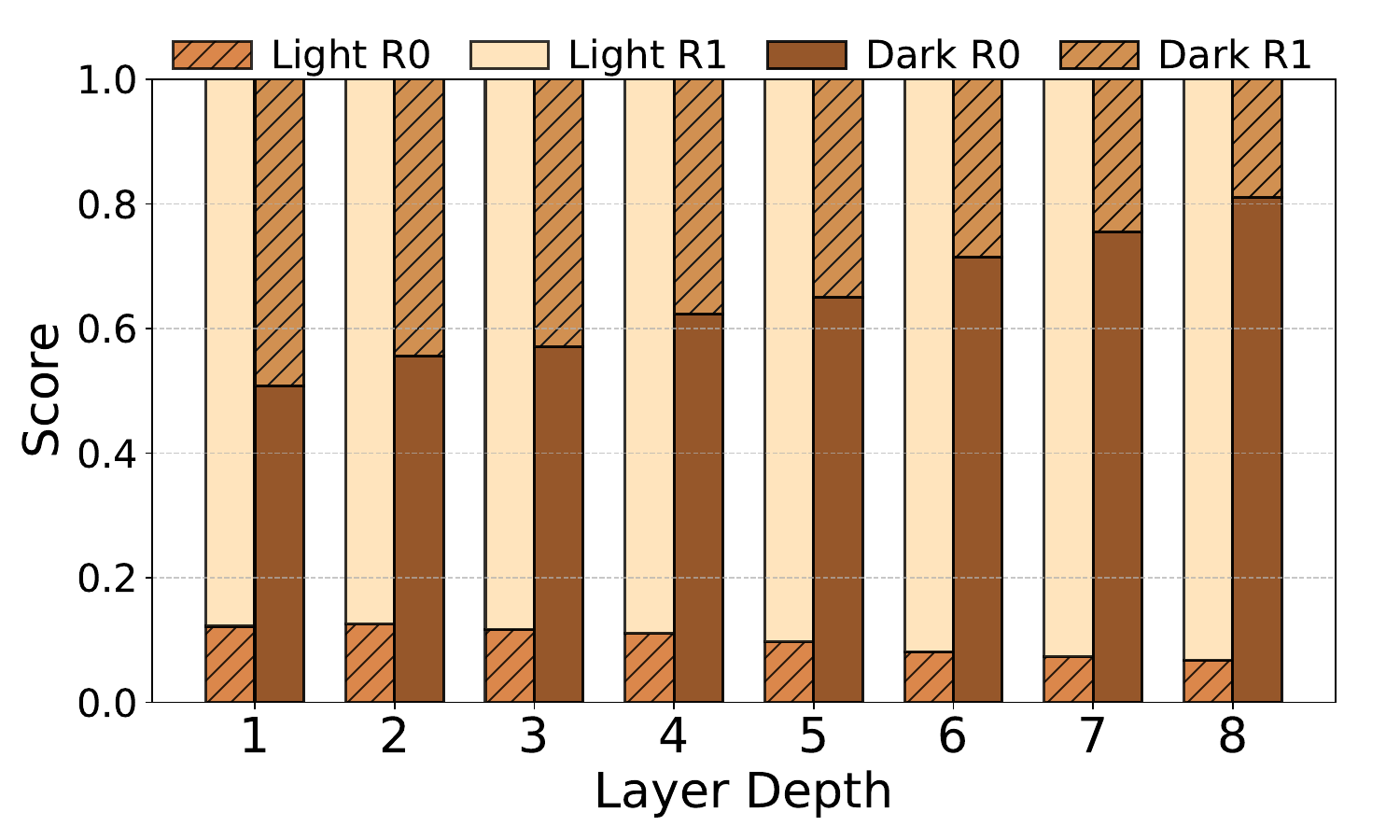}
        \caption{}
        \label{fig:router_score}
    \end{subfigure}%
    \hfill
    \begin{subfigure}[t]{0.48\textwidth}
        \centering
        \includegraphics[width=\linewidth]{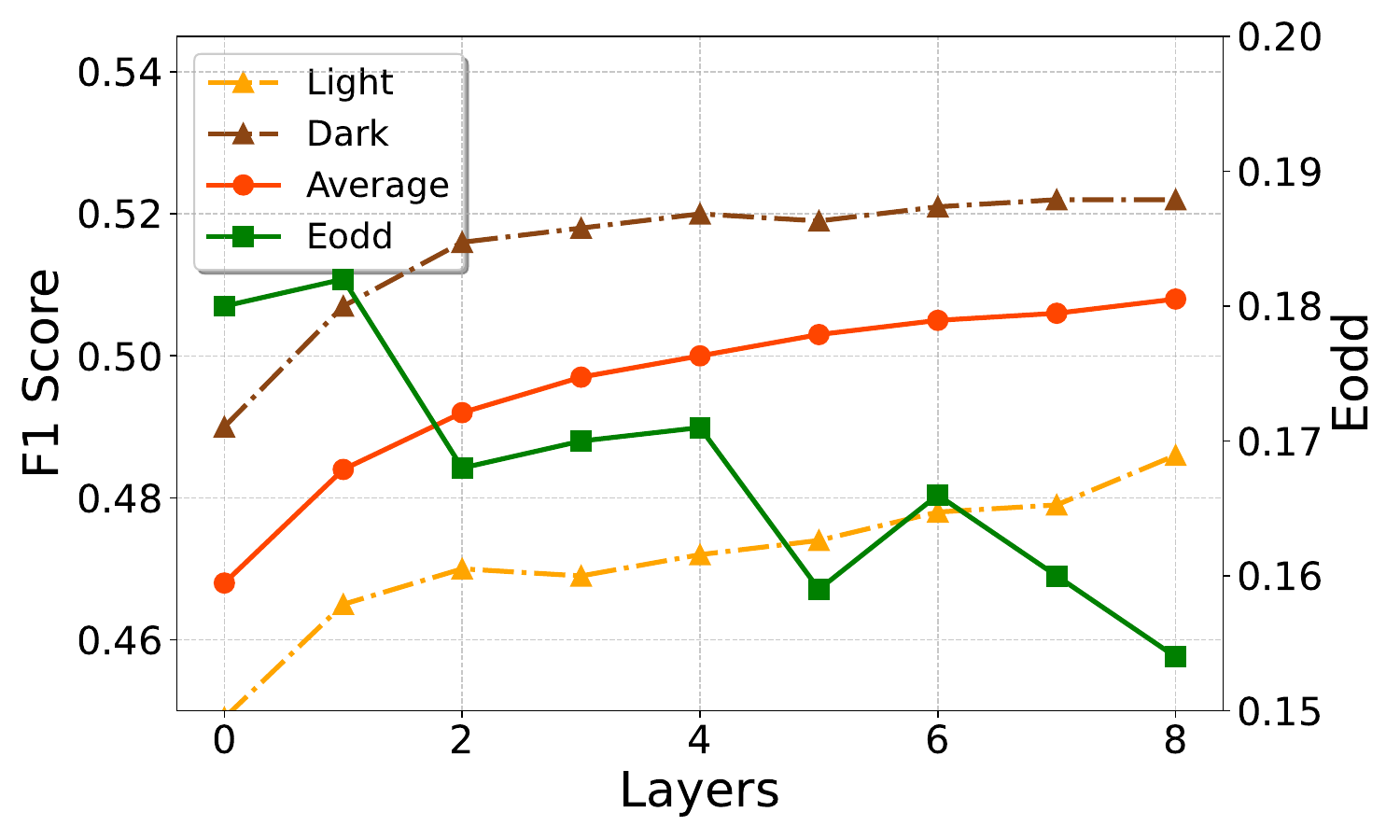}
        \caption{}
        \label{fig:ablation}
    \end{subfigure}
    \caption{(a) Router scores for dark and light skin experts across model layers. (b) Impact of the number of MoE Conv layers on model performance.}
    \label{fig:combined}
\end{figure}

\noindent\textbf{Case Study.}
In the Fitzpatrick-17k dataset, although light-skin samples outnumber dark-skin samples by nearly a 2:1 ratio, its classification performance is significantly lower for light skin. We hypothesize that dark-skin samples shift the model’s focus, negatively impacting light-skin predictions, and that training group-specific experts may better improve light-skin performance.
Figure \ref{fig:router_score} shows MoE routing scores for each skin type as network depth increases. R0 corresponds to the dark-skin expert, and R1 to the light-skin expert. 
The result shows that the tendency to select specific experts strengthens with network depth, aligning with findings in~\cite{chiu2023toward} that bias increases in deeper layers. Moreover,
the privileged group (dark skin) shows no preference for its expert in shallow layers, with selection gradually increasing in deeper layers. In contrast, the unprivileged group (light skin) strongly favors its expert from the outset, with this preference intensifying as depth increases.
These results suggest that group-specific experts may provide greater benefits to unprivileged groups.


\noindent\textbf{Ablation Study.}
To evaluate the impact of each MoE layer, we incrementally added them, starting from the final layer and gradually extending to earlier layers, as shown in Figure~\ref{fig:ablation}. The figure shows that increasing the number of MoE layers gradually improves model performance while generally improves fairness, with this effect being more pronounced in deeper layers and less significant in shallower ones.
These findings underscore a trade-off between performance, fairness, and computational resources, highlighting the flexibility of MoE layer selection based on resource availability. When computational resources are abundant, the full MoE structure can be employed for optimal performance. Conversely, in resource-constrained scenarios, MoE layers can be selectively applied to only the deeper layers, striking a balance between efficiency and effectiveness.

\section{Conclusion}

In this paper, we contend that eliminating sensitive attributes to achieve fairness is impractical in skin disease diagnosis, as these attributes provide essential diagnostic cues. Instead, we propose FairMoE, a model with layer-wise mixture-of-experts modules that incorporate group information for specialized learning. FairMoE leverages mutual information loss to establish expert-group correlations and enhances generalization through soft probabilistic routing, selectively incorporating out-of-group data. Experimental results demonstrate that FairMoE significantly improves accuracy while maintaining comparable fairness metrics.

\bibliographystyle{splncs04}
\bibliography{ref.bib}

\end{document}